%% file: clear2025.tex
\documentclass[final,12pt]{clear2025modified} % Anonymized submission

\usepackage{booktabs}
\usepackage{siunitx}
\title[DAG Learning from Zero-Inflated Count Data]{DAG Learning from Zero-Inflated Count Data Using Continuous Optimization}
\usepackage{times}
\author{%
 Noriaki Sato \\
 \normalfont{\textit{The Institute of Medical Science, The University of Tokyo}}
 \AND
 Marco Scutari \\
 \normalfont{\textit{Istituto Dalle Molle di Studi sull'Intelligenza Artificiale (IDSIA)}}
 \AND
 Shuichi Kawano \\
  \normalfont{\textit{Faculty of Mathematics, Kyushu University}}
 \AND
 Rui Yamaguchi \\
  \normalfont{\textit{Division of Cancer System Biology, Aichi Cancer Center Research Institute}} \\
  \normalfont{\textit{Division of Cancer Informatics, Nagoya University Graduate School of Medicine}}
 \AND
 Seiya Imoto \\
 \normalfont{\textit{The Institute of Medical Science, The University of Tokyo}}
}

\newcommand{\lgr}{\lambda_{\mathrm{group}}}
\newcommand{\lalign}{\lambda_{\mathrm{align}}}
\newcommand{\Paj}[1]{\mathrm{Pa#1}(j)}
\newcommand{\hldet}[1]{h_{\mathrm{ldet}}^{s}(#1)}

\begin{document}

\maketitle

\begin{abstract}%
  We address network structure learning from zero-inflated count data by casting each node as a zero-inflated generalized linear model and optimizing a smooth, score-based objective under a directed acyclic graph constraint. Our Zero-Inflated Continuous Optimization (ZICO) approach uses node-wise likelihoods with canonical links and enforces acyclicity through a differentiable surrogate constraint combined with sparsity regularization. ZICO achieves superior performance with faster runtimes on simulated data. It also performs comparably to or better than common algorithms for reverse engineering gene regulatory networks. ZICO is fully vectorized and mini-batched, enabling learning on larger variable sets with practical runtimes in a wide range of domains.%
\end{abstract}


\input{actual_content}

% \appendix

% \section{My Proof of Theorem 1}

% This is a boring technical proof.

% \section{My Proof of Theorem 2}

% This is a complete version of a proof sketched in the main text.

\end{document}

%% file: actual_content.tex
\section{Introduction}

Learning structural dependencies from observational data is a central problem in statistics and machine learning. Structural estimation methods include score-based, constraint-based, and hybrid methods that combine both \citep{schmidt_learning_2007}. In recent years, score-based continuous optimization methods such as NOTEARS, GOLEM, and DAGMA have been developed and increasingly applied to a wide range of problems \citep{nazaret_stable_2024,ng_role_2020,zheng_dags_2018,bello_dagma_2022}. These approaches transform the combinatorial problem of searching over directed acyclic graphs (DAGs) into a continuous optimization problem with differentiable constraints, thereby enabling the use of efficient gradient-based solvers and principled regularization to control model complexity.

Structure learning methods specific to count or zero-inflated count datasets, found in various domains such as medicine and biology, have also been developed. For learning DAGs from count data, \citet{park_learning_2015} proposed a method based on overdispersion scoring. \citet{choi_bayesian_2020} proposed a zero-inflated Poisson Bayesian network utilizing the Poisson distribution. \citet{choi_model_based_2023} then developed a state-of-the-art model, ZiGDAG, using hypergeometric distributions and score-based hill-climbing or Tabu greedy search algorithms. This model demonstrates robust performance on simulated data and enables recovery of biologically relevant gene regulatory networks (GRN) from single-cell transcriptomics (SCT) data. \citet{yu_directed_2023} proposed ZiDAG, a DAG learning method built upon earlier hurdle graphical models \citep{mcdavid_graphical_2019}.

These important models are useful for learning DAGs from possibly zero-inflated count data. However, the computational cost of greedy search algorithms can be substantial. In such cases, continuous optimization methods offer significant advantages due to their scalability. In this work, we develop ZICO (Zero-Inflated Continuous Optimization). This DAG learning approach incorporates the log-likelihood derived from node-wise zero-inflated negative binomial (ZINB) or Poisson (ZIP) models together with DAG constraints, using efficient optimization. Compared with state-of-the-art algorithms for structure learning from zero-inflated count data, ZICO demonstrates superior performance with faster runtimes, making it applicable to a wide range of domains.

\section{Methods}
\subsection{Preliminaries}

Bayesian networks are a class of graphical models defined over a DAG $G = (V, E)$ with vertex set $V = \{1, \ldots, d\}$ and edge set $E \subseteq V \times V$ \citep{koller}. Each vertex $j \in V$ corresponds to a random variable $X_j$ and $Pa(j)$ denotes its parents. Bayesian networks encode the conditional independencies among nodes, leading to the probability factorization
\begin{equation*}
  p(X_1, \ldots, X_d) = \prod_{j=1}^{d} p\left(X_j \mid X_{\Paj{}}\right).
\end{equation*}

\subsection{Proposed Model Architecture}

We address the DAG learning problem for zero-inflated count data by formulating the zero-inflated structural equation model underlying ZICO. Each variable $X_j, j = 1, 2, \ldots, d$ in the observational count data $X \in \mathbb{N}_0^{n \times d}$ follows a ZINB distribution conditional on its parents,
\begin{equation*}
  X_j \mid X_{\Paj{0}}, X_{\Paj{1}} \sim 
    \mathrm{ZINB}\left(\pi_j(X_{\Paj{0}}), \mu_j(X_{\Paj{1}}), r_j\right),
\end{equation*}
with link functions:
\begin{align*}
  &\mathrm{logit}\left(\pi_j(x)\right) = \gamma_j + x^T w_j^{(0)}&
  &\text{and}&
  &\log\left(\mu_j(x)\right) = \delta_j + x^T w_j^{(1)},
\end{align*}
where $w_j^{(0)}, w_j^{(1)} \in \mathbb{R}^d$ are the weights for node $j$ in the zero and count components,  
$\gamma_j, \delta_j$ are intercepts, and $r_j > 0$ is a dispersion parameter in the negative binomial (NB) distribution.  $\pi_j(X_{\Paj{0}})$ is the zero-inflation probability for variable $j$, and $\mu_j(X_{\Paj{1}})$ is the conditional mean of the NB count component for variable $j$. % 川野先生のコメントを基に追記
The resulting node-wise log-likelihood naturally separates zeros due to a structural process from the sampling zeros arising under an NB count model. Its expression for sample $i$ and node $j$ is
\begin{equation*}
  \ell_{ij} =
  \begin{cases}
    \log(1 - \pi_{ij}) + \pi_{ij} p_{ij}^{r_j} & (x_{ij} = 0) \\[6pt]
    \log \pi_{ij}
    + \log \Gamma(x_{ij} + r_j)
    - \log \Gamma(r_j)
    - \log \Gamma(x_{ij} + 1) \\
    \quad + r_j \log p_{ij}
    + x_{ij} \log(1 - p_{ij}) & (x_{ij} > 0)
  \end{cases}
\end{equation*}
where $\pi_{ij} = \mathrm{sigmoid}\left(\gamma_j + x_i^\top w_j^{(0)}\right)$, $p_{ij} = \frac{r_j}{r_j + \mu_{ij}}$, and $\mu_{ij} = \exp\left(\delta_j + x_i^\top w_j^{(1)}\right)$.

We additionally consider the ZIP model, in which each node follows a ZIP conditional distribution with the same link functions for the zero and count components:
\begin{equation*}
  X_j \mid X_{\Paj{0}},\, X_{\Paj{1}} \sim 
    \mathrm{ZIP}\bigl(\pi_j(X_{\Paj{0}}),\, \mu_j(X_{\Paj{1}})\bigr),
\end{equation*}
where $\pi_j(X_{\Paj{0}})$ is the zero-inflation probability for variable $j$ and $\mu_j(X_{\Paj{1}})$ is the conditional mean of the Poisson (or count) component for variable $j$. % 川野先生のコメントを基に追記
The node-wise ZIP log-likelihood for sample $i$ and node $j$ is
\begin{equation*}
\ell_{ij} =
  \begin{cases}
    \log\left((1 - \pi_{ij}) + \pi_{ij} e^{-\mu_{ij}}\right) & (x_{ij} = 0) \\[6pt]
    \log \pi_{ij} + x_{ij} \log \mu_{ij} \\ 
    \quad - \mu_{ij} - \log \Gamma(x_{ij} + 1) & (x_{ij} > 0)
  \end{cases}\,.
\end{equation*}

We minimize the average negative log-likelihood (NLL) for both models. We compute it fully in the log domain in both ZIP and ZINB models using the log-gamma, log-sum-exp, and softplus link functions, ensuring finite scores even when $\pi \to 1$ or $\mu \to 0$ for numerical stability \citep{cui_comprehensive_2023}.

Let $W_0 = [\, w_1^{(0)}, w_2^{(0)}, \ldots, w_d^{(0)} \,]$ and $W_1 = [\, w_1^{(1)}, w_2^{(1)}, \ldots, w_d^{(1)} \,]$ denote the coefficient matrices for the zero and count components, respectively. We enforce acyclicity on $W_0$ and $W_1$ separately. The acyclicity is imposed via a log-determinant-based function of an M-matrix transformation of the weighted adjacency matrix, introduced in DAGMA, as follows:
\begin{equation*}
  % \hldet{W} = -\log \det (s I - W \circ W) + d \log s,
  h^s_{\textrm{ldet}} (W_k) = - \log \textrm{det} (s I - W_k \circ W_k) + d \log s, \quad (k=0,1)
\end{equation*}
where $s > 0$ indicates the log-determinant parameter and $\circ$ denotes the Hadamard product.

\section{Objective Function and Continuous Optimization}

Let $\Theta = \{ W_0, W_1, \gamma, \delta, r \}$. To complete our ZICO proposal, we define the following objective with respect to $\Theta$:
\begin{multline}
  \min_{\Theta}\;
  \mu
  \left(
    -\frac{1}{n}\sum_{i=1}^{n}\sum_{j=1}^{d} \ell_{ij}
    \;+\;
    \lgr \sum_{j=1}^{d}\sum_{\substack{k=1\\ k\neq j}}^{d}
    \left\|
      \bigl( (W_0)_{kj},\, (W_1)_{kj} \bigr)
    \right\|_{2}
  \right) \\
  \;+\; \hldet{W_0} \;+\; \hldet{W_1} \;+\; \lalign \,\| W_0 - W_1 \|_{F}^{2}\,.
\label{eq:optimization}
\end{multline}

The Frobenius-norm penalty $\lalign \| W_0 - W_1 \|_{F}^{2}$ on the adjacency matrices (for both ZINB and ZIP) encourages the alignment between $W_0$ and $W_1$. Following the central-path approach used in \textsc{DAGMA}, the multiplier $\mu > 0$ is gradually decreased during training at specified epoch intervals using a decay parameter $\alpha$.

An optional elementwise group-$\ell_{1}$ penalty with nonnegative weight $\lgr\ge 0$ is applied with a cosine-annealed warm-up schedule. Specifically, the effective regularization at epoch $t$ is
\begin{equation*}
  \lambda_{\mathrm{eff}}(t) =
    \frac{\lgr}{2}
    \left(1 - \cos\bigl( \min\{1,\; t/\mathrm{warm}\}\,\pi \bigr) \right),
\end{equation*}
where $\mathrm{warm}$ controls the length of the warm-up period. This schedule gradually increases the regularization strength from $0$ to $\lgr$, dealing with early-training instability.

We use AdamW with gradient clipping \citep{adamw} for solving (\ref{eq:optimization}) to learn DAGs using ZICO. To scale to large $n$, we compute the NLL on mini-batches
$B \subseteq \{1, \ldots, n\}$ of size $|B|$:
\begin{equation*}
  \mathrm{NLL}_{B} = -\frac{1}{|B|} \sum_{i \in B}\sum_{j=1}^{d} \ell_{ij}.
\end{equation*}

By restricting training to the parameters in $W_1$ alone, the model reduces naturally to an NB or Poisson formulation (i.e., without the zero-inflation component). The optimization returns two weighted adjacency matrices $(W_0, W_1)$ in the ZINB or ZIP case, and $W_1$ in the NB or Poisson case.

\section{Experiments}

\subsection{Simulated Data}

We generated synthetic datasets with $D$ nodes and a sample size of $N = 500$. First, the true DAGs were sampled under the Erdős-Rényi (ER) and Barabasi-Albert (BA) models, with an edge probability of 0.25 for ER and expected number of three edges per node in BA \citep[Python igraph library]{csardi_igraph_2006}. 

Once the network structure was determined, we sampled the model parameters. For edges presented in the DAG, coefficients for the zero-inflation component were randomly drawn from a $U(0.5, 2)$. In contrast, coefficients for the mean component were drawn from a $U(-2, -0.5)$. In addition, intercept terms for the zero-inflation component were sampled from a $N(1.5, 0.2)$. In contrast, intercept terms for the mean component were sampled from a $N(1.5, 0.2)$. The dispersion parameter of the NB distribution was fixed at 5.0 for all nodes. % Revised in 2025/12/06 (gamma mean zero to 1.5)

Data were simulated using logic sampling to preserve the structure implied by the DAG \citep{koller}. For each node, the probability of generating a structural zero was determined from the values of its parents and the corresponding zero-inflation parameters. When a nonzero value was generated, it was sampled from a negative binomial distribution with parameters determined by the mean component of the model. This procedure produced datasets that exhibited both zero inflation and overdispersion, consistent with the topology of the specified DAG structure. For each of $D= 20$, $30$ and $50$, we generated 10 replicate datasets and learned the network structure using the methods described below. The estimated graphs were then compared with the true underlying graph using Structural Hamming Distance \citep[SHD;][]{tsamardinos_max-min_2006}, Structural Intervention Distance \citep[SID;][]{peters}, True Positive Rate (TPR), and False Discovery Rate (FDR). SHD and SID were computed with the bnlearn R package \citep{scutari_learning_2010}. For evaluation, we thresholded the absolute values of the entries of the learned matrices $W_0, W_1$ to obtain directed edges.

We compared ZICO with the state-of-the-art Greedy Equivalence Search \citep[GES; score-based, ][]{ges}, Max-Min Hill Climbing \citep[MMHC; hybrid,][]{tsamardinos_max-min_2006} and DirectLiNGAM \citep{shimizu_directlingam_2011}. As for algorithms tailored to zero-inflated data, we considered ZiDAG and ZiGDAG (score-based, hyper-Poisson and NB distributions) as well as NOTEARS with a Poisson loss (continuous optimization). For GES, MMHC, DirectLiNGAM, and ZiDAG, $\log(x+1)$-transformed count data was used as input for the structure learning. For the other algorithms, the raw simulated count data were used as the input. We used the implementation of GES in the pcalg R package \citep{markus_kalisch_causal_2012}, and those of DirectLiNGAM and MMHC in bnlearn. The equivalence class learned by GES was replaced by its consistent DAG extension before computing SHD and SID. In this experiment, the epochs, $\alpha$, and the first $\mu$ were set to 4000, 0.1, and 1, respectively. Weighted adjacency matrices were binarized at the absolute value threshold of 0.3.

The performance of various algorithms for recovering DAGs using the simulated zero-inflated data summarized in Table~\ref{tab:perf-D50}. Our ZICO proposal demonstrated comparable or superior accuracy to other approaches, with faster execution times than ZiGDAG and ZiDAG. The speed-up was most pronounced when the number of variables was large.
As for structural accuracy, ZICO also yielded the lowest errors, achieved in the ZINB model with $D = 50$ and in the ER model, indicating close reconstruction of the true networks.

The superior performance of the methods based on ZINB and ZIP is expected, as these methods explicitly model the zero-inflated nature of the data. By aligning the statistical assumptions with the underlying data-generating process, they achieve higher accuracy compared with conventional approaches.

% Used conversion library
% Revised on 2025/12/07, 2025/12/11
\begin{table}[t]
  \centering
  \small
  \scriptsize
  \setlength{\tabcolsep}{3pt}
  % Use resizebox to fit the table
  \resizebox{\linewidth}{!}{%
    \begin{tabular}{
      l
      % S[table-format=1.3] % TPR
      % S[table-format=1.3] % FDR
      % S[table-format=5.3] % Time (s)
      % S[table-format=3.1] % SHD
      % S[table-format=4.1] % SID
      c % TPR
      c % FDR
      c % Time (s)
      c % SHD
      c % SID
      l                   % Graph model
    }
    \toprule
    {Algorithm} &
    \multicolumn{1}{c}{TPR} &
    \multicolumn{1}{c}{FDR} &
    \multicolumn{1}{c}{Time (s)} &
    \multicolumn{1}{c}{SHD} &
    \multicolumn{1}{c}{SID} &
    {Graph model} \\
    \midrule
    DirectLiNGAM       & 0.002 (0.002)        & 0.965 (0.040)        & 108.480 (9.788)          & 313.500 (18.222)         & 2282.600 (101.654)        & ER \\
    GES                & 0.076 (0.019)        & 0.759 (0.059)        & \textbf{0.500 (0.081)}   & 318.100 (21.471)         & 2566.300 (122.864)        & ER \\
    MMHC               & 0.081 (0.016)        & 0.671 (0.060)        & 1.287 (0.368)            & 314.100 (17.754)         & 2380.900 (135.685)        & ER \\
    NOTEARS            & 0.116 (0.029)        & \textbf{0.106 (0.057)} & 72.144 (68.673)        & 287.900 (16.603)         & 2127.100 (98.747)         & ER \\
    ZICO (NB)      & 0.758 (0.031)        & 0.479 (0.029)        & 48.239 (38.484)          & 229.500 (21.824)         & 1831.200 (195.047)        & ER \\
    ZICO (Poisson) & 0.590 (0.053)        & 0.595 (0.038)        & 42.975 (37.211)          & 276.700 (32.582)         & 2251.700 (154.873)        & ER \\
    ZICO (ZINB)    & \textbf{0.780 (0.020)} & 0.407 (0.020)      & 55.643 (39.179)          & \textbf{180.500 (17.116)} & \textbf{1660.100 (164.593)} & ER \\
    ZICO (ZIP)     & 0.748 (0.042)        & 0.421 (0.037)        & 51.975 (38.649)          & 188.500 (28.321)         & 1792.800 (201.923)        & ER \\
    ZiDAG              & 0.016 (0.009)        & 0.507 (0.223)        & 789.730 (90.073)         & 303.300 (16.269)         & 2222.600 (94.768)         & ER \\
    ZiGDAG             & 0.100 (0.013)        & 0.590 (0.041)        & 21571.112 (1560.249)     & 283.200 (18.432)         & 2474.100 (118.722)        & ER \\
    ZiGDAG (NB)        & 0.102 (0.018)        & 0.589 (0.057)        & 1869.899 (343.458)       & 282.500 (19.086)         & 2456.200 (89.170)         & ER \\
    \hline
    DirectLiNGAM       & 0.022 (0.024)        & 0.930 (0.062)        & 81.601 (0.972)           & 150.800 (2.394)          & 2435.000 (55.102)         & BA \\
    GES                & 0.365 (0.057)        & 0.580 (0.056)        & \textbf{0.634 (0.097)}   & 125.000 (10.853)         & 2257.600 (135.318)        & BA \\
    MMHC               & 0.306 (0.034)        & 0.543 (0.044)        & 11.419 (8.484)           & 129.400 (8.959)          & 2328.400 (82.141)         & BA \\
    NOTEARS            & 0.284 (0.066)        & \textbf{0.220 (0.093)} & 118.902 (111.393)      & 124.000 (7.087)          & 2398.200 (95.381)         & BA \\
    ZICO (NB)      & 0.791 (0.048)        & 0.506 (0.043)        & 41.966 (33.133)          & 137.300 (18.815)         & \textbf{1212.900 (220.299)} & BA \\
    ZICO (Poisson) & 0.658 (0.065)        & 0.495 (0.049)        & 41.041 (35.064)          & 114.100 (10.837)         & 1638.100 (279.596)        & BA \\
    ZICO (ZINB)    & \textbf{0.792 (0.038)} & 0.321 (0.036)      & 48.842 (33.577)          & \textbf{74.300 (8.001)}  & 1246.800 (219.674)        & BA \\
    ZICO (ZIP)     & 0.760 (0.038)        & 0.321 (0.036)        & 47.724 (35.264)          & 76.500 (9.698)           & 1376.100 (223.746)        & BA \\
    ZiDAG              & 0.045 (0.030)        & 0.609 (0.145)        & 990.282 (132.780)        & 139.800 (3.765)          & 2451.900 (31.593)         & BA \\
    ZiGDAG             & 0.402 (0.059)        & 0.511 (0.048)        & 18428.821 (4741.086)     & 102.300 (7.558)          & 2264.100 (115.928)        & BA \\
    ZiGDAG (NB)        & 0.413 (0.041)        & 0.494 (0.043)        & 1086.489 (683.056)       & 100.600 (6.381)          & 2237.900 (119.419)        & BA \\
      \bottomrule
      \end{tabular}
  }
  \caption{Performance comparison ($D = 50$). Values are averages over 10 replicates. The values inside parentheses are standard deviations. The best performing values are shown in bold. HP, hyper-Poisson; NB, negative binomial; ZICO: our proposal.}
  \label{tab:perf-D50}
\end{table}

\subsection{Sign and Norm-Comparison Experiment}

To assess the sensitivity of ZICO to different data-generating processes, we evaluated the sign patterns between $W_0$ and $W_1$. For each graph type $g\in\{\mathrm{BA},\mathrm{ER}\}$ and dimension $D = 20$, we generated a random DAG $B$ as described previously, fixed $N=500$ observations, and drew edge weights from uniform ranges under different sign configurations.

We experimented with four sign configurations:
\begin{equation*}
  \left(\mathrm{sign}(W_0), \mathrm{sign}(W_1)\right) \in \left\{ (+,+),\,(-,-),\,(+,-),\,(-,+) \right\}.
\end{equation*}
Unless otherwise stated, $(+,-)$ denotes $W_0 > 0$ (zero link) and $W_1 < 0$ (count link), and $(-,+)$ denotes the opposite. For example, in $(+,+)$, parents make zeros more likely, but conditional on being nonzero, counts are larger. This yields many zeros with occasionally large positive counts.

We also compare three ways of coupling $W_0$ and $W_1$: 1) alignment with the Frobenius norm, at $\lalign \in \{0, 0.1, 1\}$; 2) alignment with an elementwise $\ell_{1}$ norm, at the same $\lalign$; 3) a combined approach that aggregates the two matrices via elementwise $\ell_{2}$ pooling $\left(\widetilde{W} = \sqrt{W_0^{\circ2} + W_1^{\circ2} + \varepsilon}\right)$ and enforces acyclicity on $\widetilde{W}$. Furthermore, we vary $\lgr \in \{0, 0.001, 0.01 \}$.

We evaluate couplings using the area under the precision-recall curve (AUPRC) for the combined $W_0$ and $W_1$, based on the edge ranking obtained from the weighted adjacency matrix. We generated five data replicates for each experimental setting.

The results of the experiments are summarized in Table~\ref{tab:best_W0W1_AUPRC}. The optimal tuning parameter values for ZICO varied across the four sign configurations. Still, we found that the separate acyclicity plus light $\ell_{1}$ alignment ($\lalign = 0.1$) when the $W_0$ is positive, yields the best AUPRC. Heavy alignment and large group penalties consistently underperformed. Notably, the $(-,+)$ combination performed markedly worse than other choices across all experimental settings. We considered the reason to be a mismatch between data generation and likelihood, and the improvements there likely require changes beyond the model architecture or hyperparameters rather than stronger regularization.

% Revised on 2025/12/07, 2025/12/11
\begin{table}[t]
  \centering
  \small
\begin{tabular}{rrrrr rrrrl r}
\toprule
Configuration & $W_0low$ & $W_0high$ & $W_1low$ & $W_1high$ & $g$ & $\lgr$ & $\lalign$ & Norm & AUPRC \\
  \midrule
$(-,-)$ & -2.0 & -0.5 & -2.0 & -0.5 & BA & 0.010 & 0.0 & none     & 0.826 (0.079) \\
$(-,-)$ & -2.0 & -0.5 & -2.0 & -0.5 & ER & 0.010 & 0.0 & none     & 0.812 (0.065) \\
$(-,+)$ & -2.0 & -0.5 &  0.5 &  2.0 & BA & 0.000 & 0.0 & none     & 0.325 (0.041) \\
$(-,+)$ & -2.0 & -0.5 &  0.5 &  2.0 & ER & 0.000 & 0.0 & none     & 0.460 (0.125) \\
$(+,-)$ & 0.5 &  2.0 & -2.0 & -0.5 & BA & 0.001 & 0.1 & $\ell_1$ & 0.772 (0.091) \\
$(+,-)$ & 0.5 &  2.0 & -2.0 & -0.5 & ER & 0.010 & 0.1 & $\ell_1$ & 0.797 (0.043) \\
$(+,+)$ & 0.5 &  2.0 &  0.5 &  2.0 & BA & 0.001 & 0.1 & $\ell_1$ & 0.858 (0.059) \\
$(+,+)$ & 0.5 &  2.0 &  0.5 &  2.0 & ER & 0.000 & 0.1 & $\ell_1$ & 0.865 (0.042) \\
\bottomrule
\end{tabular}
\caption{Best ZICO parameter combinations for each $(W_0, W_1, g)$ group maximizing AUPRC ($W_0W_1$). $W_0low$, $W_0high$, $W_1low$, $W_1high$ indicate the lower and upper bounds of uniform distribution in $W_0$ and $W_1$. The values inside parentheses are standard deviations.}
\label{tab:best_W0W1_AUPRC}
\end{table}

\subsection{Simulations for different support for $W_0$ and $W_1$}

To decouple patterns between the zero-inflation and count components and characterize the effect of the alignment penalty term, we partitioned the edge set of the DAG into two support masks with a user-specified overlap fraction~$\rho$, yielding $\mathbf{M}_0$ for the zero-inflation mechanism and $\mathbf{M}_1$ for the mean model. Edge weights $W_0$ and $W_1$ were sampled on their respective supports, $\mathbf{M}_0$ and $\mathbf{M}_1$. As zero-inflation and count processes may not hold in all contexts, we varied $\rho \in \{0, 0.25, 0.5, 0.75, 1\}$ to simulate how the proposed algorithm performs under different alignment penalties and $\lalign$. We evaluated $g \in \{\mathrm{BA},\mathrm{ER}\}$ and $(+,+)$ for generating data under the ZINB model in this experiment. We generated five data replicates for each condition.

\begin{figure}[t]
  \centering
  \includegraphics[width=\linewidth]{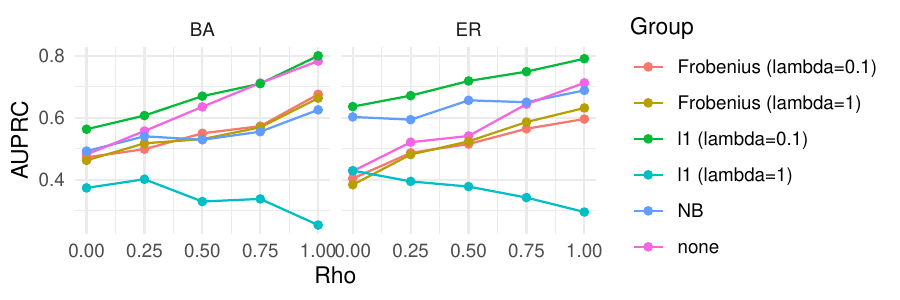}
  \caption{The AUPRC across multiple overlap levels $\rho$. At higher overlap, the no alignment setting achieves performance comparable to the alignment settings, whereas at lower overlap, introducing a light alignment term can improve performance. Note that AUPRC for $W_1$ is reported for NB model.}
  \label{fig:ovl}
\end{figure}

Under $(+,+)$, the alignment strength directly applied to the chosen norm yields a $\rho$-dependent trade-off. When the overlap is low, a light $\ell_{1}$ alignment is the best choice. In contrast, no alignment with the coupled acyclicity rose steadily with $\rho$ and performed comparable to the alignment setting at high overlap in our experiments, indicating that when the common backbone between $W_0$ and $W_1$ is strong, forcing acyclicity to coupled adjacency matrix can be beneficial, yielding comparable performance in the $(+,+)$ case under the BA model (Figure~\ref{fig:ovl}).

\subsection{Simulated SCT Data}

Bayesian networks have been widely applied to GRN inference from transcriptomics data obtained from microarray and RNA-seq technology \citep{imoto_combining_2003}. We conducted an evaluation using scMultiSim, a state-of-the-art SCT data simulator that supports GRN simulations \citep{li_scmultisim_2025}. 

We first generated random DAGs under the BA model with the expected number of three neighbours per node, and assigned effect sizes to each edge by sampling from a $U(1, 5)$. We then generated cell-count data with $N = 500$ cells, trained networks on them  with ZICO, evaluated the learned DAGs accuracy. As a means of comparison, we also performed GRN inference using GENIE3 \citep{huynh-thu_inferring_2010}, GRNBoost2 \citep{moerman_grnboost2_2019}, LEAP \citep{specht_leap_2017}, SINCERITIES \citep{papili_gao_sincerities_2018}, NOTEARS \citep{zheng_dags_2018}. For GENIE3, GRNBoost2, and ZICO, we used the raw count data. For the other data, a $\log(x+1)$ transformation was applied before the inference. We used L2 loss in NOTEARS inference. Pseudo-time ordering, which is necessary for some algorithms, was calculated using the slingshot R package \citep{street_slingshot_2018}.
As most GRN inference methods do not explicitly produce a DAG, accuracy was evaluated using the AUPRC ratio, defined as the ratio of the AUPRC value to that obtained when using random edges.

Whether the individual molecular identifier count data are zero-inflated is a matter of debate \citep{cao_umi_2021}. This can be modelled naturally by omitting the $W_0$ term in (\ref{eq:optimization}), thus reducing it to an NB or Poisson regression NLL. Therefore, in this experiment, we used NB, Poisson, ZINB, and ZIP models, and computed AUPRC using the absolute values of the $W_1$ coefficient matrix in the NB and Poisson models and the combined $W_0$ and $W_1$ in the ZINB and ZIP models. We chose the alignment penalty to be the $\ell_{1}$ norm with $\lalign = 0.1$, which performed best in earlier $(+,-)$ experiments where the zero link and the count link act in opposite directions. In SCT data, such a configuration is consistent with the empirical behavior of dropout stochastic detection failures, which are more likely at low true expression, so that higher dropout co-occurs with lower observed positive counts \citep{kharchenko_bayesian_2014}.

\begin{figure}[t]
  \centering
  \includegraphics[width=\linewidth]{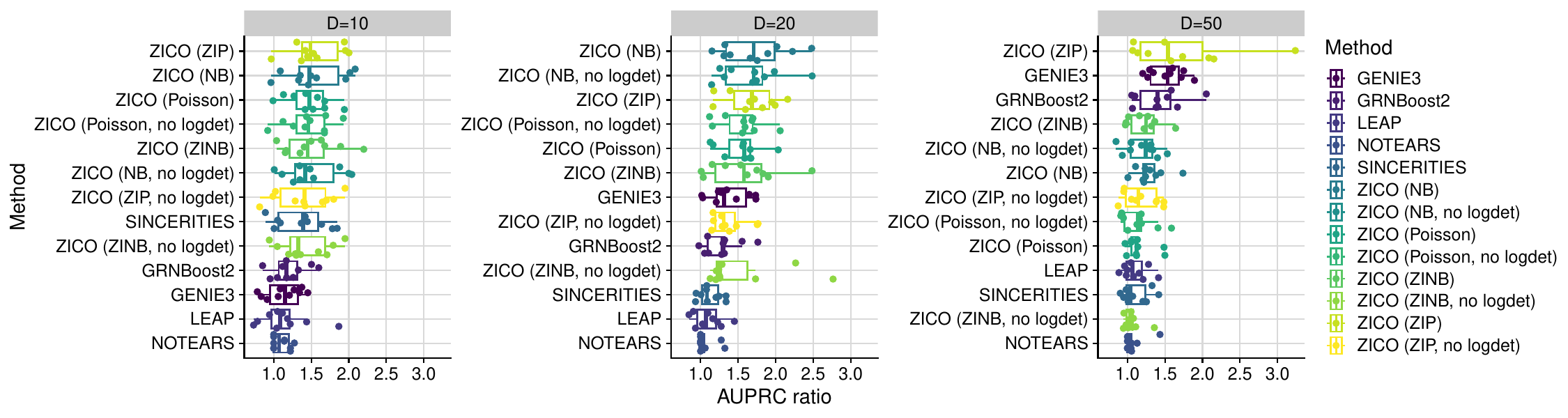}
  \caption{Performance in recovering transcriptomic regulatory relationships, assessed by the AUPRC ratio (\(\mathrm{AUPRC}/\mathrm{AUPRC}_{\text{random}}\)). Algorithms are ordered by the median values.}
  \label{fig:auprc}
\end{figure}

The results of reverse engineering GRNs using ZICO are shown in Figure~\ref{fig:auprc} along with those from other commonly used directed GRN inference methods. Overall, the assessment using the AUPRC ratio relative to random prediction shows that the performance in recovering GRN is generally low, consistent with previous studies. ZICO achieved results comparable to or better than other commonly used methods for GRN inference. 
We hypothesize that the reason is that ZICO provides a close surrogate for the Poisson–Beta models used in the scMultiSim. DAG constraints, which are absent from other GRN inference algorithms, could also enhance performance; the ablation of the log-determinant term degraded performance in some settings.

In addition, we performed dropout simulation experiments on the clean count data.  Following previous studies \citep[e.g.,][]{dibaeinia_sergio_2020}, we introduced expression-dependent dropout into the clean data. We then evaluated ZICO with both ZINB/ZIP and NB/Poisson models on the resulting datasets, comparing their performance using the same AUPRC-ratio metric.

Specifically, we applied a probabilistic dropout to an input count matrix $X \in \mathbb{N}_0^{n \times d}$ and returned $Y = X \circ B$. First, we performed the elementwise transformation $z_{ij}=\log(x_{ij}+1)$ and pooled $\{z_{ij}\}$ across all entries. Let $m=\mathrm{Percentile}_q(\{z_{ij}\})$ be the $q$-th percentile that defines the threshold of a logistic function. For each entry, the retention probability is $p_{ij}=\{1+\exp[-\alpha(z_{ij}-m)]\}^{-1}$ with slope $\alpha>0$; larger $x_{ij}$ yield larger $z_{ij}$ and hence larger $p_{ij}$. We then sampled an independent Bernoulli mask $b_{ij}\sim\mathrm{Bernoulli}(p_{ij})$ for all $(i,j)$ and obtained the dropout-perturbed output by $Y=X\circ B$ with $B=(b_{ij})$. The percentile $q$ controls the location of the threshold. We used $\alpha = 1$ and $q = 65$ in the experiment. % Hadamard product

\begin{figure}[t]
  \centering
  \includegraphics[width=\linewidth]{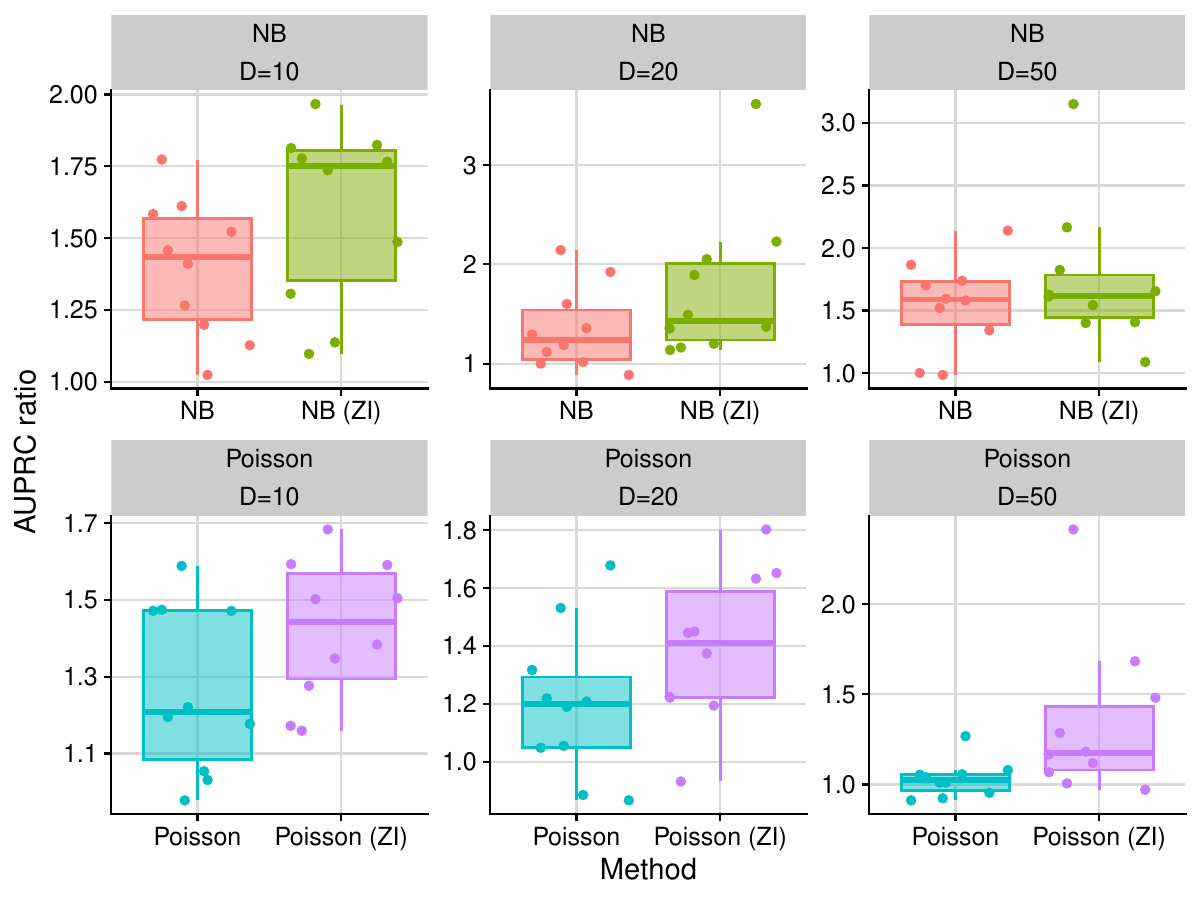}
  \caption{Comparison of the performance of recovering GRN from the count data applied dropout simulation. The boxplot comparing ZINB, ZIP, NB, and Poisson models assessed by the AUPRC ratio (\(\mathrm{AUPRC}/\mathrm{AUPRC}_{\text{random}}\)) is shown.}
  \label{fig:dropout}
\end{figure}

The results are shown in Figure~\ref{fig:dropout}. The ZINB and ZIP approaches, which incorporate $W_0$, consistently achieved higher AUPRC ratios across all $D$ values than the NB and Poisson models. This suggests that estimators that account for zero inflation are advantageous when estimating DAG from SCT data with dropout events.

% I refer to the following SERGIO function:
% def dropout_indicator(self, scData, shape = 1, percentile = 65):
%     """
%     This is similar to Splat package

%     Input:
%     scData can be the output of simulator or any refined version of it
%     (e.g. with technical noise)

%     shape: the shape of the logistic function

%     percentile: the mid-point of logistic functions is set to the given percentile
%     of the input scData

%     returns: np.array containing binary indactors showing dropouts
%     """
%     scData = np.array(scData)
%     scData_log = np.log(np.add(scData,1))
%     log_mid_point = np.percentile(scData_log, percentile)
%     prob_ber = np.true_divide (1, 1 + np.exp( -1*shape * (scData_log - log_mid_point) ))

%     binary_ind = np.random.binomial( n = 1, p = prob_ber)

%     return binary_ind

\subsection{Evaluation Environment}

The proposed ZICO is implemented in PyTorch. All code is available from \url{http://www.github.com/noriakis/ZICO}. Training was carried out on the supercomputer SHIROKANE using PyTorch 2.6.0 \citep{paszke_pytorch_2019}. We used ggplot2 for data visualization. The performance statistics were calculated using scikit-learn \citep{scikit-learn}. The supercomputing resource was provided by the Human Genome Centre, the Institute of Medical Science, the University of Tokyo.

% Below is annonimized version
% The proposed ZICO is implemented in PyTorch. All code is available from \url{https://osf.io/4whak/overview?view_only=5016c8f0893c48588f0c9f6e197dd831}. Training was carried out on a high-performance computing cluster using PyTorch 2.6.0 \citep{paszke_pytorch_2019}. We used ggplot2 for data visualization, and performance statistics were calculated using scikit-learn \citep{scikit-learn}.

\section{Conclusions}

We developed ZICO, an accurate and efficient algorithm to learn DAGs from zero-inflated count data through continuous optimization. Compared to existing methods, ZICO scales to larger numbers of variables, making it practical in real-world settings such as GRN inference from SCT data.

Continuous optimization methods have been criticized for issues with varsortability \citep{reisach_beware_2021}. Because the variance is not estimated independently from the mean but determined by the underlying distributional form, this issue should not be relevant in this setting. Nevertheless, a theoretical analysis of identifiability and potential biases under different data-generating mechanisms remains an important direction for future work.

Reverse engineering GRN from SCT data has been reported to be challenging in several works \citep{chen_evaluating_2018,dibaeinia_sergio_2020}. Zero-inflation is also still debated in SCT data; the ZIP/ZINB components in the model help with clearly zero-inflated data. They are otherwise over-parameterized \citep{cao_umi_2021}, as we have shown using simulated SCT data: NB models may then outperform zero-inflated models. In practice, this suggests that model choice should account for both empirical characteristics of the data and prior knowledge of the underlying measurement technology. ZICO can naturally accommodate such choices, since the same optimization scheme can be coupled with different count distributions, including NB and Poisson variants. As a limitation, our empirical evaluation focused on explicitly acyclic structures; feedback loops and other non-DAG regulatory motifs, which are common in biology, and are not captured by the current formulation.

% Clear 2026 Acknowledgments---Will not appear in anonymized version
% For submission of clear, delete section and insert \acks
% \acks{This study is partially supported by Takeda Science Foundation, JSPS KAKENHI 25K21331; and grant K25-2170 from the International Joint Usage/Research Center, the Institute of Medical Science, the University of Tokyo.}

% This part is for preprint.
\section{Acknowledgements}
This study is partially supported by Takeda Science Foundation, JSPS KAKENHI 25K21331; and grant K25-2170 from the International Joint Usage/Research Center, the Institute of Medical Science, the University of Tokyo.
\bibliography{ref}